# Agentività e telicità in GilBERTo: implicazioni cognitive

**Agnese Lombardi, Alessandro Lenci**
Università di Pisa, Pisa, Italia
`a.lombardi19@studenti.unipi.it, alessandro.lenci@unipi.it`

## Abstract

**English.** The goal of this study is to investigate whether a Transformer-based neural language model infers lexical semantics and use this information for the completion of morphosyntactic patterns. The semantic properties considered are telicity (also combined with definiteness) and agentivity. Both act at the interface between semantics and morphosyntax: they are semantically determined and syntactically encoded. The tasks were submitted to both the computational model and a group of Italian native speakers. The comparison between the two groups of data allows us to investigate to what extent neural language models capture significant aspects of human semantic competence.

**Italiano.** *L'obiettivo di questo studio è quello di indagare se neural language models basati su Transformer inferiscono aspetti semantico-lessicali rilevanti per l'interfaccia con la sintassi ed utilizzano queste informazioni per il completamento di task morfosintattici. Le proprietà semantiche considerate sono la telicità (anche in relazione all'individuazione) e l'agentività. Entrambe sono semanticamente determinate e sintatticamente codificate. I task sono stati sottoposti sia al modello che ai parlanti. La comparazione tra i due gruppi di dati ci permetterà di determinare se questi modelli computazionali catturano aspetti significativi della competenza semantica umana.*

## 1 Introduzione

L'ipotesi distribuzionale stabilisce che lessemi con contesti linguistici simili hanno un significato simile (Wittgenstein, 1953; Harris, 1954; Firth, 1957).

I modelli distribuzionali sono stati impiegati con successo in molti task di Natural Language Processing, ma quale siano le conoscenze acquisite durante il processo di addestramento rimane una questione ancora aperta.

Uno degli approcci per comprendere la natura di queste informazioni linguistiche consiste nel valutare la loro accuratezza in task psicolinguistici[1]. Alcuni studi hanno indagato proprietà e dipendenze sintattiche (Linzen et al., 2016; Ettinger, 2016; Wilcox et al., 2018; Futrell et al., 2019; Marvin e Linzen, 2018; Hu et al., 2020; Lau et al., 2020), altri si sono concentrati su aspetti semantici e pragmatici come: la similarità (Hill et al.,2015), la categorizzazione (Baroni e Lenci, 2010), l'analogia (Mikolov et al., 2013), la negazione (Marvin e Linzen, 2018; Jumelet e Hupkes, 2018), il ragionamento pragmatico, i ruoli semantici e la conoscenza eventiva (Ettinger, 2020).

Il nostro lavoro contribuisce a questa linea di ricerca e ne adotta l'approccio psicolinguistico, ma se ne discosta nelle proprietà investigate, proponendo l'analisi della telicità (in combinazione con l'individuazione) e dell'agentività.

L'obiettivo è indagare se l'inferenza di queste proprietà semantico-lessicali favorisce l'elaborazione di alcuni task morfosintattici. Nella nostra analisi abbiamo scelto di utilizzare un modello distribuzionale di tipo predittivo che utilizza rappresentazioni contestualizzate (Peters et al., 2018; Devlin et al., 2019): GilBERTo, un modello distribuzionale italiano ispirato all'architettura di RoBERTa (Liu et al., 2019).

## 2 Aspetti della semantica di interfaccia: azionalità e agentività

Un importante dominio dell'informazione lessicale riguarda l'evento e i suoi partecipanti. Se da un lato l'aspetto è una nozione di natura eminentemente morfologica e semantica, che riguarda le modalità di svolgimento dell'evento (piuttosto che la sua localizzazione e la serie di rapporti temporali); l'*azionalità verbale*, d'altra parte, non viene codificata dalla morfologia flessiva. Non basta dire che l'azionalità sia un fatto inerente al significato intrinseco di un lessema, bisogna individuare delle classi coerenti di verbi, contraddistinte da un comportamento sintattico omogeneo nell'ambito della lingua considerata. Ci sono diversi aspetti lessicali che possono essere codificati da una classe verbale e che pertengono alla classificazione azionale.

---

[1] Gli stimoli, nei task psicolinguistici, sono progettati in modo da fornire informazioni sulle proprietà linguistiche che influenzano il comportamento umano (giudizio grammaticale, velocità di letture o risposte neurali).

Vendler (1967) categorizza le classi azionali sulla base di tre proprietà fondamentali (la duratività, la dinamicità e la telicità) e individua quattro gruppi principali: i verbi stativi (*states*), di attività (*activities*), risultativi (*accomplishments*) e trasformativi (*achievements*). I verbi trasformativi e quelli risultativi vengono raggruppati nella categoria dei *telici*. Gli eventi telici hanno la caratteristica di essere finalizzati al raggiungimento di un telos, ovvero una meta o una fine.

### 2.1 Telicità

Finora abbiamo considerato la classe dei telici come l'insieme dei risultativi e dei trasformativi. In ogni caso, è necessario specificare che la telicità può essere intesa come un continuum, un asse semantico che vede ai due estremi i prototipi della categoria (inerentemente telici e inerentemente atelici) e al centro gli elementi che definiremo *configurazionali*[2]. Dunque, la telicità non è una proprietà discreta e non sempre è possibile definirla in maniera inequivocabile (non essendo determinata solo dai tratti lessicali): essendo fortemente dipendente dal contesto (dagli argomenti del verbo e dalla transitività, ma anche dalla coniugazione dell'aspetto verbale) e veicolata dal senso complessivo della frase. Ad esempio, "disegnare" e "cantare" sono di per sé predicati verbali non telici: ciò che li rende telici, in un determinato contesto, è la presenza di un oggetto diretto che li determina, finalizzandoli al raggiungimento di un preciso scopo.

In "Gennaro ha disegnato/ha cantato tutto il pomeriggio" il predicato verbale si configura come atelico, ma diventa telico se si ha "Gennaro ha disegnato il ritratto di mia nonna" o "Gennaro ha cantato la sua canzone preferita".

Da Bertinetto (1997), apprendiamo che un test per distinguere l'accezione telica di un verbo da quella non-telica è l'aggiunta dell'avverbiale "in x tempo"[3] che risulta incompatibile con i predicati non telici. L'applicazione dell'avverbiale "per X tempo", invece, risulta o incompatibile con i predicati verbali telici o se compatibile, ne neutralizza la telicità.

### 2.2 Telicità e individuazione

L'individuazione dell'oggetto o del soggetto può incidere sull'interpretazione telica assegnata all'evento. Il concetto di individuazione unifica diverse proprietà dell'argomento e può essere considerata anch'essa una proprietà semantica di interfaccia, perché agisce sia a livello semantico che morfosintattico. L'individuazione si riferisce alla propensione di un'entità ad essere concepita come un individuo indipendente. Possiamo considerare l'individuazione come la risultante delle seguenti proprietà: individui, animatezza, concretezza/astrattezza, singolare/plurale, mass/count, referenziale/non-referenziale (Romagno, 2005). Concependo l'individuazione come un continuum, i significati possono essere raggruppati secondo classi di equivalenza che condividono le stesse proprietà di individuazione e le classi di individuazione possono essere ordinate sulla base del loro grado di individuazione. Il grado di individuazione di un'entità può essere calcolato sulla base della media derivata tramite l'unione dei valori di tutti i fattori che la determinano. Considereremo [+ individuato] un argomento umano, proprio, animato, concreto, singolare, numerabile, referenziale e [- individuato] un argomento inanimato, comune, astratto, plurale, non numerabile, non referenziale. Sia dal punto di vista semantico, sia da quello morfosintattico, si registra un'influenza reciproca tra individuazione e telicità. Ad esempio, nella frase "mangiare del pane", l'oggetto è poco individuato; nella frase "mangiare una pagnotta di pane", invece, l'oggetto è individuato e rappresenta l'argomento interno diretto del predicato. Ne consegue che nel primo caso l'interpretazione assegnata all'evento è atelica e nel secondo caso è telica.

### 2.3 Agentività

Secondo Cruse (1971) l'agentività è presente in ogni frase che si riferisce ad un'azione effettuata

---

[2] Con questo termine faremo riferimento, d'ora in poi, a quei predicati verbali la cui interpretazione telica (o atelica) è determinata dal contesto (in particolare dall'individuazione dell'oggetto o del soggetto).

[3] "X tempo" simboleggia un'espressione temporale numericamente quantificata: in due minuti, in due giorni, in due ore, in due anni…

da un'entità che impiega la propria energia per condurre l'azione. Nella definizione di entità sono inclusi gli esseri viventi, alcuni tipi di macchine ed eventi naturali. Da ciò è possibile dedurre che l'argomento agentivo è prototipicamente il soggetto, essendo esso il promotore dell'azione, ed è sempre associato con una struttura logica[4] di attività; e che solo verbi che possiedono nella loro struttura logica un predicato di attività possono avere un argomento agentivo. Nella struttura logica di un predicato, l'agentività è rappresentata come DO (x, [do (x, ...]. Ad esempio, se si confrontano i verbi "kill" e "murder" (il primo verbo può accogliere soggetti inanimati, mentre il secondo no) la struttura logica si configura come: kill: [do (x, Ø)] CAUSE [BECOME dead (y)] murder: DO (x, [do (x, Ø] CAUSE [BECOME dead (y)]) (Pustejovsky e Batiukova, 2019).

Ci sono altri verbi, però, che possono assumere un'interpretazione agentiva. Infatti, il più delle volte, l'agentività è determinata dal modo in cui un verbo è utilizzato all'interno di una frase e non è un'inerente proprietà lessicale del verbo. In questi casi, l'agentività non fa parte del significato lessicale del verbo e non è rappresentata nella sua struttura logica, piuttosto è determinata da implicazioni basate sull'animatezza dell'attore e sulle proprietà lessicali del verbo. Holisky (1987) sostiene che l'interpretazione agentiva spesso sorge dall'intersezione tra le proprietà semantiche all'interno di una frase (le proprietà semantiche dell'attore NP e del predicato) e i principi generali di conversazione.

Un test molto semplice per capire se l'agentività è lessicalizzata in un verbo coinvolge l'avverbio "inavvertitamente" e consiste nel verificare se il suo impiego crea una contraddizione all'interno della frase. Se la frase diventa contraddittoria, allora il predicato verbale lessicalizza l'agentività.

È il caso di: "Gennaro ha assassinato *inavvertitamente il suo vicino", in cui la contraddizione è evidente, quindi il predicato è agentivo. Anche l'agentività, come la telicità, è una proprietà che agisce nell'interfaccia tra sintassi e semantica.

## 3 Esperimento

Il nostro obiettivo è, quindi, indagare se GilBERTo è in grado di inferire la telicità (anche in combinazione con l'individuazione) e l'agentività e di utilizzare questa inferenza per il completamento di task morfosintattici. Inoltre, vogliamo determinare se l'elaborazione del modello può essere comparata a quella dei parlanti nei medesimi task.

Essendo entrambe queste proprietà semantiche, codificate morfosintatticamente, possiamo determinarne la corretta elaborazione mediante dei test morfosintattici. La risposta selezionata sarà dunque informativa dal punto di vista semantico.

Per garantire un confronto diretto tra il modello e i parlanti, ad entrambi verranno sottoposti i medesimi task.

### 3.1 Stimoli

I soggetti e il modello dovevano completare dei cloze test con la giusta opzione morfosintattica.

Abbiamo ideato tre task. Il primo task indaga la telicità, il secondo l'individuazione in rapporto alla telicità e il terzo l'agentività. Ogni task è composto da sessanta frasi affermative con verbo coniugato al passato prossimo.

Nel primo task sulla telicità le frasi dovevano essere completate con la preposizione "in" o "per" nelle locuzioni avverbiali "in/per X tempo". I soggetti sono nomi comuni, impiegati alla terza persona, animati e, a volte, utilizzati con il supporto di un aggettivo possessivo. Abbiamo incluso verbi inerentemente telici (sia risultativi che trasformativi), verbi inerentemente atelici e verbi configurazionali (20 + 20 + 20). Nelle seguenti frasi, estratte dal primo task, riportiamo esempi con verbo telico (1), atelico (2) e configurazionale (3):

*(1) L'operaio ha demolito la casa in/per un'ora*
*(2) Mia sorella ha dormito in/per tre ore*
*(3) Il ragazzo ha corso in/per un'ora*

Nel secondo task, che indaga la telicità in relazione all'individuazione, abbiamo utilizzato lo stesso cloze test. Tuttavia, abbiamo strutturato un design fattoriale che divide le frasi in quattro gruppi (di 15 frasi), secondo lo schema seguente:

I gruppo: soggetto [+ ind][5] e oggetto [- ind]
II gruppo: soggetto [+ ind] e oggetto [+ ind]
III gruppo: soggetto [- ind] e oggetto [- ind]
IV gruppo: soggetto [- ind] e oggetto [+ ind][6]

---

[4] "[…] Logical Structures (LS) consisting of constants, which mostly represent predicates, and modifiers (BECOME, INGR, CAUSE, etc.). […] these elements are not words from any natural language, but items of a semantic metalanguage" (Van Valin e LaPolla, 1997).
[5] Ind = individuato

[6] Per i soggetti [+ individuati] abbiamo utilizzato nomi comuni di persona con aggettivi possessivi; per i soggetti [- individuati] nomi comuni plurali o nomi astratti. Gli oggetti [- individuati] sono costituiti da nomi comuni (riferiti a liquidi, plurali o nomi massa) con un aggettivo qualificativo senza articolo determinativo

In ogni gruppo abbiamo incluso predicati verbali telici, atelici e configurazionali (5+5+5). Riportiamo una frase per ognuno dei quattro gruppi:

*I Mio fratello ha bevuto latte fresco in/per cinque minuti*

*II Mio fratello ha bevuto un bicchiere di latte in/per cinque minuti*

*III I mobili hanno accumulato della polvere densa in/per dieci anni*

*IV I mobili hanno accumulato un sacco di polvere in/per dieci anni*

Nel terzo task, che indaga l'agentività, le frasi dovevano essere completate con "inavvertitamente" o "intenzionalmente". Abbiamo variato sia le proprietà del soggetto (includendo soggetti con il ruolo prototipico di actor, ma anche soggetti meno prototipici) e quelle dell'oggetto (includendo oggetti con il ruolo prototipico di undergoer, ma anche oggetti meno prototipici). Abbiamo incluso predicati verbali che hanno la proprietà dell'agentività lessicalizzata nella loro struttura semantica (quindi inerentemente agentivi), predicati verbali inerentemente inagentivi e predicati verbali che possono assumere entrambi i valori a seconda del contesto (20 + 20 + 20). Anche in questo caso abbiamo escluso i nomi propri ed i soggetti sono tutti animati ed alla terza persona. Il seguente esempio riporta rispettivamente un verbo agentivo, inagentivo e configurazionale:

*(4) Mio fratello ha deciso intenzionalmente/inavvertitamente di scegliere*

*(5) Mio fratello è invecchiato intenzionalmente/inavvertitamente*

*(6) Mio padre ha cotto intenzionalmente/inavvertitamente per molto tempo la carne*

Nei primi due task, le frasi sottoposte al modello, contenevano una parola mascherata [7] nell'input e il modello doveva fornire come output, al suo posto, le prime cinque opzioni più probabili e le relative probabilità. Nel terzo task invece, il modello fornisce come output direttamente la frase completa con una delle opzioni. I parlanti, invece, dovevano scegliere in ognuno dei tre task l'opzione preferibile tra le due proposte.

### 3.2 Partecipanti

65 volontari madrelingua italiani avevano il compito di completare le frasi scegliendo l'opzione più opportuna. Ai parlanti venivano fornite le istruzioni per il completamento dei task all'inizio degli stessi. Tutti i dati sono stati raccolti tramite Google Forms.

### 3.3 Modello

GilBERTo è un modello del linguaggio italiano preaddestrato basato sull'architettura di RoBERTa e sull'approccio di tokenizzazione del testo di CamemBERT. Il modello è stato addestrato con la tecnica di mascheramento delle subwords per 100k passi gestendo 71 GB di testo italiano con 11.250.012.896 parole (OSCAR: Open Super-large Crawled Almanach coRpus). È stato considerato un vocabolario di 32k BPE (Byte-Pair Encoding) subwords, generate usando SentencePiece tokenizer. Nei primi due task è stata utilizzata la libreria pytorch\fairseq Python e nel terzo task la libreria FitBERT.

## 4 Risultati

Le teorie linguistiche stabiliscono che i verbi inerentemente telici dovrebbero selezionare "in x tempo", mentre gli inerentemente atelici "per x tempo". I verbi configurazionali selezionano "in" o "per" a seconda dell'interpretazione telica che il soggetto vuole conferire alla frase. I dati del primo task confermano questo schema, come illustra la tabella 1, in cui sono riportate le preferenze del modello[8] e dei parlanti.

| Predicati | Modello (%) | | Parlanti (%) | |
|---|---|---|---|---|
| | in | per | in | per |
| Telici | 70 | 20 | 100 | 0 |
| Atelici | 80 | 10 | 0 | 100 |
| Config. | 35 | 30 | 50 | 50 |

Table 1: Primo task

Come si evince, i dati dei verbi inerentemente telici e inerentemente atelici di entrambi i gruppi presentano pochissima dispersione: l'interpretazione è telica o atelica a seconda del predicato verbale e non c'è indecisione tra le opzioni proposte.

---

("latte fresco"); con gli oggetti [+ individuati] si assiste allo schema opposto: nomi al singolare, con articolo determinativo, o nomi leggeri quantificatori ("un sacco di polvere"). Per favorire una comparazione diretta tra due frasi abbiamo utilizzato lo stesso predicato verbale e lo stesso soggetto (nei primi due gruppi [+ individuato] e negli ultimi due [- individuato]) e abbiamo variato solo l'individuazione dell'oggetto.

[7] Ad esempio: *Il ragazzo ha corso <mask> un'ora.*

[8] Le percentuali del modello, nelle tabelle 1 e 2, corrispondono al valore della mediana, calcolata in relazione alle probabilità fornite dal modello come output.

I verbi configurazionali invece, presentano, sia nel modello che nei parlanti, una dispersione dei dati più ampia e nessuna delle due opzioni risulta preferibile.

I dati del secondo task, raccolti nella tabella 2, presentano uno scenario più complesso.

| Predicati | Modello (%) | | Parlanti (%) | |
|---|---|---|---|---|
| | in | per | in | per |
| Telici (I gruppo) | 60 | 15 | 45 | 55 |
| Telici (II gruppo) | 35 | 10 | 100 | 0 |
| Telici (III gruppo) | 60 | 15 | 80 | 20 |
| Telici (IV gruppo) | 30 | 20 | 100 | 0 |
| Atelici (I gruppo) | 0 | 80 | 0 | 100 |
| Atelici (II gruppo) | 20 | 80 | 80 | 20 |
| Atelici (III gruppo) | 10 | 60 | 40 | 60 |
| Atelici (IV gruppo) | 15 | 55 | 75 | 25 |
| Config. (I gruppo) | 0 | 90 | 0 | 100 |
| Config. (II gruppo) | 10 | 70 | 100 | 0 |
| Config. (III gruppo) | 20 | 40 | 30 | 70 |
| Config. (IV gruppo) | 20 | 60 | 40 | 60 |

Table 2: Secondo task

I dati mostrano che il modello seleziona "in" solo con i verbi inerentemente telici (si registra uno scarto maggiore tra le due opzioni nel primo e nel terzo gruppo, in cui l'oggetto è [- individuato]). I parlanti, invece, con verbi inerentemente telici selezionano "in" in ognuno dei gruppi, tranne che nel primo (soggetto [+ individuato] e oggetto [- individuato]), in cui "per" viene preferito nel 55% dei casi. Contrariamente al modello, in cui, nel secondo e nel quarto gruppo (con oggetti [+ individuati]), "in" ottiene una probabilità vicina a quella di "per", nei parlanti è del 100% ("per" non viene mai selezionato).

Con verbi inerentemente atelici, invece, i parlanti selezionano "in" quando l'oggetto è [+ individuato] e "per" quando l'oggetto è [- individuato] (nel secondo e nel quarto gruppo, quindi, rispettivamente nell'80% e nel 75% dei casi). Nel modello, invece, l'opzione "per" risulta preferibile in ognuno dei quattro casi considerati. Infine, con verbi configurazionali i parlanti mostrano una preferenza per "in" nel secondo gruppo (soggetto e oggetto [+ individuati]) e per "per" nei restanti tre. Nello specifico, però, nel primo gruppo (soggetto [+ individuato] e oggetto [- individuato]) "per" risulta vincente nel 100% dei casi (confermando i dati dei verbi inerentemente telici, in cui, nel primo gruppo, i parlanti selezionavano "per" al 55%); mentre, nel terzo e nel quarto gruppo (con soggetti [- individuati]), "per" ottiene percentuali più basse, determinando conseguentemente uno scarto inferiore tra le due opzioni. Il modello rispetta lo schema dei parlanti con la variazione delle probabilità di "per" tra primo gruppo (con soggetto [+ individuato] ha un valore del 90%) e terzo e quarto (con soggetto [-individuato] ha rispettivamente il 40% e il 60%); ma non conferma lo schema dei parlanti nel secondo gruppo (ha il 70%, mentre dai parlanti non veniva mai scelto).

Nella tabella 3 sono raccolti i dati[9] del terzo task.

| Predicati | Modello (%) | Parlanti (%) |
|---|---|---|
| Agentivi | inavv. (70) | intenz. (100) |
| Inagentivi | inavv. (65) | inavv. (100) |
| Config. | inavv. (60) | inavv. (50) |

Table 3: Terzo task

I risultati del terzo task mostrano che il modello sceglie l'opzione "inavvertitamente" in più dl 50% delle frasi, per ognuno dei tre gruppi di verbi, nonostante la variazione di agentività. I parlanti, invece, mostrano coerenza con le ipotesi linguistiche.

## 5 Discussione

L'analisi aveva lo scopo di testare l'elaborazione della telicità (anche in rapporto all'individuazione) e dell'agentività, sia nei parlanti che nel modello, e di indagare se quest'elaborazione determina il giusto completamento morfosintattico. I dati mostrano che i parlanti operano in maniera coerente con l'ipotesi proposta e con la teoria linguistica. Infatti, è la telicità a determinare la giusta codifica morfosintattica. Inoltre, mostrano un'influenza evidente dell'individuazione sull'interpretazione telica. In presenza di verbi inerentemente telici i parlanti selezionano "in" senza essere influenzati dalla minore individuazione del soggetto. L'unico caso in cui la valenza telica inerente del predicato subisce una variazione è con soggetto [+ individuato] e oggetto [- individuato]. Questo stesso schema non si riscontra con soggetto [- individuato] e oggetto [- individuato].

Sappiamo che l'oggetto riveste prototipicamente il ruolo di paziente e che quindi è colui che nel caso di un evento telico subisce il mutamento di stato, quindi [+ coinvolto] e [+ individuato][10]. D'altra parte, il soggetto è prototipicamente il promotore dell'azione, quindi [- coinvolto] e [- individuato] dell'oggetto. In questo caso, possiamo supporre che a guidare l'interpretazione atelica (nonostante la telicità inerente del predicato) sia la non prototipicità dei due argomenti nella frase[11]. A riprova, ciò non si verifica nelle frasi del terzo gruppo, in cui non vi è differenza tra coinvolgimento ed individuazione del soggetto e dell'oggetto.

---

[9] Le percentuali del modello, nella tabella 3, corrispondono al numero di frasi in cui il modello ha preferito inavvertitamente o intenzionalmente.

Anche i dati comportamentali dei verbi inerentemente atelici risultano coerenti con l'ipotesi: i parlanti conferiscono un'interpretazione telica alle frasi in cui l'oggetto è [+ individuato] e atelica a quelle in cui è [- individuato]. Con i verbi configurazionali l'interpretazione telica è possibile solo se entrambi gli argomenti sono individuati. I dati comportamentali, inoltre, riflettono la natura scalare della telicità: i verbi configurazionali sono quelli che riportano una dispersione dei dati più ampia.

Questo tipo di elaborazione si riscontra anche nel modello, dove i verbi configurazionali non mostrano nessuna preferenza netta a favore di una delle due opzioni, dimostrando che il modello riesce ad inferire la natura scalare della telicità. Tuttavia, una differenza emerge nell'elaborazione dell'individuazione. Da un lato i parlanti sono influenzati dall'individuazione nell'interpretazione telica; d'altra parte, il modello non mostra la stessa sensibilità. Questa mancata elaborazione dell'individuazione viene confermata dal fatto che con verbi inerentemente telici il modello predilige sempre un'interpretazione telica e con verbi inerentemente atelici un'interpretazione atelica. Quindi l'accordo tra modello e parlanti è determinato dalle proprietà del predicato e non dall'individuazione di soggetto e oggetto.

Anche per l'agentività i parlanti mostrano coerenza con l'ipotesi e con le teorie linguistiche. Nei casi in cui l'agentività è codificata nell'informazione eventiva del verbo viene selezionato con una preferenza netta l'avverbio ad essa associato. Viceversa, accade con verbi inerentemente inagentivi; mentre i verbi configurazionali non mostrano preferenza per nessuna delle due opzioni. Il modello, al contrario, completa il task senza essere influenzato dall'agentività. Questo risultato potrebbe essere determinato dal tipo di task o dall'utilizzo di FitBERT, che per la prima volta viene applicato ad un modello basato sull'architettura di RoBERTa.

Generalizzando, il modello riesce ad utilizzare le proprietà semantiche veicolate dal predicato per determinare la giusta codifica morfosintattica: elabora, quindi, la telicità in coerenza con le teorie linguistiche, come una proprietà scalare. Tuttavia, non si può affermare che questa elaborazione avvenga anche per le proprietà semantiche che sono veicolate dal contesto dell'intera frase: l'agentività o la variazione della telicità dovuta all'individuazione.

## 6 Conclusioni

La differenza di elaborazione tra modello e parlanti ci permette di proporre delle implicazioni dal punto di vista teorico. La prima implicazione è che seppure questi modelli mostrino una certa sensibilità e una certa aderenza al modo in cui i parlanti processano il linguaggio, non possono essere considerati un modello cognitivo di elaborazione del linguaggio. Tuttavia, questa analisi ci permette di ipotizzare la codifica di queste proprietà di semantica lessicale nell'informazione vettoriale dei modelli distribuzionali, facendo luce su quali sono le informazioni semantiche codificate.

Sicuramente esiste un'influenza distribuzionale nel modo in cui i parlanti utilizzano le informazioni, ma bisogna considerare anche fattori che dipendono dal contesto extralinguistico. In lavori futuri ha senso continuare ad indagare l'elaborazione di proprietà di semantica lessicale nei modelli distribuzionali, magari adottando altre tecniche di indagine e confrontando dati estratti da modelli diversi. Futuri lavori potranno indagare altre proprietà di semantica lessicale: ad esempio l'intransitività scissa.

Inoltre, il nostro lavoro può essere migliorato includendo lo studio dell'aspetto verbale e dell'influenza che questo ha nell'interpretazione della frase (coniugando le frasi non solo all'aspetto perfettivo, ma anche all'imperfettivo). Ad esempio, potrebbe essere interessante considerare il caso delle lingue slave che grammaticalizzano la telicità attraverso l'opposizione tra aspetto perfettivo e imperfettivo. Infine, questi studi potrebbero essere utilizzati per implementare questi modelli distribuzionali, migliorando il modo in cui veicolano la composizionalità semantica (a livello frasale).

---

[10] Telicità, coinvolgimento e individuazione dell'oggetto sono anche alcuni dei parametri che determinano la transitività di una frase.

[11] Il soggetto, in questo caso, ha le proprietà semantiche prototipiche dell'oggetto e, quest'ultimo, ha quelle prototipiche del soggetto.

# References

Nikos Athanasiou, Elias Iosif and Alexandros Potamianos. 2018. Neural activation semantic models: compositional lexical semantic models of localized neural activations. In *Proceedings of the 27th International Conference on Computational Linguistics*, 2867–2878, Association for Computational Linguistics, Santa Fe, New Mexico, USA.

Marco Baroni and Alessandro Lenci 2010. Distributional Memory: A General Framework for Corpus-Based Semantics. In *Computational Linguistics*, 36: 673–721.

Pier Marco Bertinetto. 1986. Tempo, aspetto e azione nel verbo italiano, Firenze, Accademia della Crusca.

Lucia Busso, Ludovica Pannitto e Alessandro Lenci. 2018. Modelling the Meaning of Argument Constructions with Distributional Semantics. Are constructions enough? In *Proceedings of the Fifth Italian Conference on Computational Linguistics CLiC-it 2018*.

Alan Cruse. 1973. Some Thoughts on Agentivity. In *Journal of Linguistics*, 9(1):11-23.

Jacob Devlin et al. 2018. BERT: Pre-training of Deep Bidirectional Transformers for Language Understanding. In *CoRR, Vol. abs/1810.04805*.

Guy Emerson. 2020. What are the Goals of Distributional Semantics? In *Proceedings of the 58th Annual Meeting of the Association for Computational Linguistics*, 1-18.

Allyson Ettinger. 2019. What {BERT} is not: Lessons from a new suite of psycholinguistic diagnostics for language models. In *CoRR, Vol. abs/1907.13528*.

Cécile Fabre and Alessandro Lenci. 2015. Distributional Semantics Today. Introduction to the special issue. In *Traitement automatique des langues. Sémantique distributionelle (ATALA)*, 50(2) :7-20.

Scott Grimm. 2018. Grammatical number and the scale of individuation. In *Language*, 94:527-594.

Alessandro Lenci e Alessandra Zarcone. 2008. Computational Models of Event Type Classification in Context. In *Language Resources and Evaluation Conference*.

Alessandro Lenci. 2008. Distributional semantics in linguistic and cognitive research. *In The Italian Journal of Linguistics*, 20:1-20.

Yang Liu et al. 2019. RoBERTa: A Robustly Optimized BERT Pretraining Approach, In *CoRR, Vol. abs/1907.11692*.

Timothee Mickus et al. 2020. What do you mean, BERT? Assessing BERT as a Distributional Semantics Model. In *Proceedings of the Society for Computation in Linguistics*, 3(34):1-12.

Giulio Ravasio and Leonardo Di Perna. GilBERTo: An Italian pretrained language model based on RoBERTa. DOI: https://github.com/idb-ita/GilBERTo.

Vered Shwartz and Ido Dagan. 2019. Still a Pain in the Neck: Evaluating Text Representations on Lexical Composition. *In CoRR, Vol. abs/1902.10618*.

Ian Tenney et al. 2019. What do you learn from context? Probing for sentence structure in contextualized word representations. In *CoRR, Vol. abs/1905.06316*.

Peter D. Turney. 2006. Expressing Implicit Semantic Relations without Supervision. In *CoRR, Vol abs/cs/0607120*.

Robert D. van Valin and Randy J. LaPolla. 1997. Syntax: Structure, Meaning, and Function. In *Linguistics, Cambridge University Press*.

Zeno Vendler. 1967. Causal relations. In *Journal of Philosophy*, 64:704-713.

Koki Washio and Tsuneaki Kato. 2018. Neural Latent Relational Analysis to capture lexical semantic relation in a vector space. In *Proceedings of the 2018 Conference on Empirical Methods in Natural Language Processing,* 594-600.

Mo Yu and Mark Dredze. 2014. Improving Lexical Embeddings with Semantic Knowledge. In *Proceedings of the 52nd Annual Meeting of the Association for Computational Linguistics*, 2:545-550.

Xunjie Zhu, Tingfeng Li and Gerard de Melo. 2018. Exploring Semantic Properties of Sentence Embeddings, In *Proceedings of the 56th Annual Meeting of the Association for Computational Linguistics*, 2:632-637.